%% file: acl_latex.tex
\pdfoutput=1

\documentclass[11pt]{article}

\usepackage[preprint]{acl}

\usepackage{times}
\usepackage{latexsym}

\usepackage[T1]{fontenc}

\usepackage[utf8]{inputenc}

\usepackage{microtype}

\usepackage{inconsolata}

\usepackage{graphicx}
\usepackage{amssymb}

\usepackage{algorithm} %
\usepackage[noend]{algpseudocode}

\usepackage{tabularx}
\usepackage{multirow}
\usepackage{arydshln}

\usepackage[most]{tcolorbox}
\usepackage{xcolor, colortbl}
\definecolor{bggray}{rgb}{1,1,1}
\newtcolorbox[list inside=prompt,auto counter]{prompt}[1][]{
    colbacktitle=black!10,
    fonttitle=\small,
    coltitle=black,
    colback=white,
    fontupper=\footnotesize,
    boxsep=4pt,
    left=0pt,
    right=0pt,
    top=0pt,
    bottom=0pt,
    boxrule=1pt,
    #1,
}

\newcommand\extrafootertext[1]{%
    \bgroup
    \renewcommand\thefootnote{\fnsymbol{footnote}}%
    \renewcommand\thempfootnote{\fnsymbol{mpfootnote}}%
    \footnotetext[0]{#1}%
    \egroup
}

\newcommand{\sss}{{\texttt{AMRS}$^\texttt{\scriptsize{3}}$}~}

\title{Semantic Graphs for Syntactic Simplification:\linebreak A Revisit from the Age of LLM}

\author{Peiran Yao \and Kostyantyn Guzhva \and Denilson Barbosa\\
Department of Computing Science\\
University of Alberta\\
{\tt \{peiran, denilson\}@ualberta.ca}}

\begin{document}
\maketitle
\begin{abstract}
  Symbolic sentence meaning representations, such as AMR (Abstract Meaning Representation) provide expressive and structured semantic graphs that act as 
  intermediates that simplify downstream NLP tasks.
  However, the instruction-following capability of large language models (LLMs) offers a shortcut to effectively solve NLP tasks, questioning the utility of semantic graphs.
  Meanwhile, recent work has also shown the difficulty of using meaning representations merely as a helpful auxiliary for LLMs.
  We revisit the position of semantic graphs in syntactic simplification, the task of simplifying sentence structures while preserving their meaning, which requires semantic understanding, and evaluate it on a new complex and natural dataset.
  The AMR-based method that we propose, {\texttt{AMRS}$^\texttt{\scriptsize{3}}$}, demonstrates that state-of-the-art meaning representations can lead to easy-to-implement 
  simplification methods with competitive performance and unique advantages in cost, interpretability, and generalization.
  With \sss as an anchor, we discover that syntactic simplification is a task where semantic graphs are helpful in LLM prompting.
  We propose AMRCoC prompting that guides LLMs to emulate graph algorithms for explicit symbolic reasoning on AMR graphs, and show its potential for improving LLM on semantic-centered tasks like syntactic simplification.
\end{abstract}

\input{intro.tex}
\input{related_work.tex}

\input{task.tex}
\input{amrs3.tex}
\input{prompting.tex}

\input{misc.tex}

\bibliography{custom}

\end{document}

%% file: intro.tex
\section{Introduction}
\label{sec:intro}

\extrafootertext{\vspace{-3mm}\\To appear at TextGraphs-17 @ ACL 2024. Code, models, and data are available at \url{https://github.com/U-Alberta/AMRS3}.}

Frameworks for symbolic sentence meaning representations, exemplified by UCCA \cite{abend-rappoport-2013-ucca}, Abstract Meaning Representation (AMR) \cite{banarescu-etal-2013-abstract}, and UMR \cite{vangysel2021umr}, provide varying levels of abstraction away from the lexical and syntactical structures of natural language sentences, commonly in the form of \textit{semantic graphs} \cite{oepen-etal-2020-mrp}.
Compared to dense representations such as semantically meaningful embeddings \cite{reimers-gurevych-2019-sentence},
representing the meaning of a sentence as a graph allows for the use of classical (and explainable) algorithms (e.g. traversal and partition) to ease the development of more controllable and interpretable methods for semantic-focused NLP applications, including but not limited to 
text simplification \cite{sulem-etal-2018-simple}, question answering from knowledge bases \cite{kapanipathi-etal-2021-leveraging}, and text-style transfer \cite{shi-etal-2023-amr}.

Meanwhile, large language models (LLMs), representatively the ChatGPT \cite{ouyang2022instruct,openai2023gpt4} and Llama \cite{llama3modelcard} families, have demonstrated prevailing performance in the aforementioned applications.
Their instruction following capability \cite{ouyang2022instruct} enables training-free adaptation to specific tasks, which, in terms of the burden for implementation, is at a similar level to that of writing graph algorithms on top of semantic graphs.
This prompts researchers to rethink the role of symbolic meaning representations in the era of LLMs, and to explore the potential of combining the two paradigms, with the negative findings that directly appending AMR to the input of LLMs is not beneficial, if not harmful, in many tasks \cite{jin2024analyzing}.

Along these lines, we study the task of syntactic simplification and aim to answer two research questions: 
\textbf{RQ1} (\S\ref{sec:amrsss}): Can state-of-the-art meaning representing semantic graphs provide a light-weight, easy-to-implement, and interpretable alternative to LLMs for this task?
\textbf{RQ2} (\S\ref{sec:prompting}): Can it be helpful to supply semantic graphs as auxiliaries to LLMs to improve their performance on this task? 

Syntactic simplification, including variants like Split and Rephrase \cite{narayan-etal-2017-split}, sentence splitting \cite{niklaus-etal-2019-minwikisplit} and \citet{gao-etal-2021-abcd}, is a type of text simplification task that aims to rewrite sentences to reduce the syntactic complexity while preserving its meaning, typically operationalized by converting a complex text into a set of atomic sentences with simpler structures.
It has practical applications in improving text accessibility for less-proficient readers \cite{watanable2009facilita}, improving weaker NLP pipelines \cite{niklaus2023dissim}, and detecting hallucination in complex statements \cite{hou2024probabilisticframeworkllmhallucination}.
Despite modifying syntactic structures as the outcome, the task is inherently semantic-focused, as sentences are expected to be atomic in meaning and semantically equivalent to the original complex sentence, making semantic graphs a natural choice as an intermediate.

To answer RQ1, we propose \sss (shorthand for \textbf{A}bstract \textbf{M}eaning \textbf{R}epresentation for \textbf{S}yntactic \textbf{S}entence \textbf{S}implification), 
a simple yet effective graph-based algorithm that breaks down the AMR graph of a complex sentence into a set of subgraphs, each corresponding to a semantic unit. The subgraphs then guide the generation of simpler sentences which form the final output.
AMR is chosen as it is the meaning representation that receives more attention in recent developments of treebanks \cite{ldc2020amr}, parsing \cite{xu2023symbolllm}, text generation \cite{bai2022amrbart}, and cross-lingual adaptation \cite{10.1162/coli_a_00503},
and it reflects the state-of-the-art of graph-based meaning representation.
We demonstrate that with a well-developed semantic graph like AMR, a syntactic simplification system can be derived from simple rules as a lightweight alternative to LLMs.
Evaluations on the synthetic WebSplit \cite{narayan-etal-2017-split} dataset and real-world complex sentences from a Humanities corpus \cite{orlando-2023} show that \sss yields simplifications that are comparable to those of complex existing systems and LLMs in terms of both syntactic simplicity and meaning preservation, while enjoying, in principle, the merits of simplicity, interpretability, and language-neutrality.

It is unsurprising that LLM outperforms symbolic methods in syntactic simplification \cite{ponce2023split}.
We aim to answer RQ2 and see whether AMR still has merits as an auxiliary to LLMs (namely GPT-3.5 and Llama-3-8B) in this task.
Contrary to \citeposs{jin2024analyzing} report that directly adding AMR to the input is harmful in many tasks, we find syntactic simplification slightly benefits from the auxiliary AMR inputs.
We investigate what elements of AMR are helpful to LLMs in our case, and find that prompting in Chain-of-Code \cite{li2023chainofcode} style allows LLMs to emulate the execution process of \sss and perform reasoning over AMR graphs, providing insights on how AMR can be made a useful auxiliary for LLMs in this and other semantic-centered tasks.

We contribute a LLM-era's perspective on graphical approaches toward the long-standing task of syntactic simplification:
the task is benchmarked on a hard and natural complex sentence dataset that we construct;
we offer a reference point of the latest developments in symbolic meaning representations for the task;
and finally, we provide insights on the role of symbolic meaning representations in the era of LLMs that complement recent work.

%% file: related_work.tex
\section{Related Work}
\label{sec:related_work}
\paragraph*{Text Simplification.}
Syntactic simplification is a subtask of automated text simplification, the problem of improving text readability and understandability while retaining information, that has a wide spectrum of forms \cite{althanyyan2022automated}: complementing syntactic simplification, lexical simplification focuses on replacing complex words with simpler synonyms \cite{paetzold2017survey}.
Meanwhile, summarization is another form of simplification that removes superfluous information or unnecessary details \cite{nenkova2011automatic}.
Given the difference in focuses, general text simplification benchmarks and evaluations such as those of \citet{maddela-etal-2023-lens} and \citet{alvamanchego2021unsuitability} do not directly apply to syntactic simplification in isolation.

\paragraph*{Syntactic Simplification.}
Prior to LLMs, syntactic simplification was commonly modeled as a sequence-to-sequence task where systems are trained on parallel corpora synthesized from knowledge graphs \cite{narayan-etal-2017-split}, mined from Wikipedia \cite{botha-etal-2018-learning} and translations \cite{kim-etal-2021-bisect}, or crowd-sourced \cite{gao-etal-2021-abcd}.
These specialized models struggle to generalize to unseen data, which our work demonstrates is solvable with simple rule-based methods combined with a powerful semantic representation (AMR).
This combination is admittedly not a new idea: DisSim \cite{niklaus2023dissim} is a performative simplification system, yet it relies on a larger set of expert-crafted lexical rules that is not as simple and transferrable as our approach.
DSS \cite{sulem-etal-2018-simple} uses UCCA as the semantic representation, and we inherit its idea and build on AMR which is more powerful.
\citet{ponce2023split} evaluates fine-tuning LLMs on a split-and-rephrase dataset, while our analysis on LLM focuses on the zero-shot instruction-following setting.

\paragraph*{Symbolic Reasoning for LLM.}
\citet{jin2024analyzing} suggest that adding serialized AMR graphs to the input of LLM in a direct manner is not effective in prompting LLM to perform implicit reasoning over the AMR graph. This is consistent with the observation that LLM needs guidance on task decomposition to perform complex reasoning \cite{wei2022cot} such as manipulating AMR.
However, symbolic data, such as code and AMR, likely has the potential to benefit LLM \cite{yang2024llmcode}.
Our work investigates whether methods prompting LLM to perform explicit symbolic reasoning, such as Chain-of-Code \cite{li2023chainofcode}, can be more helpful than direct prompting as in \citet{jin2024analyzing}.
An alternative to prompting, which is beyond the scope of this work, is to fine-tune the LLM across symbolic reasoning tasks including AMR to improve its reasoning ability \citet{xu2023symbolllm}.

%% file: task.tex
\begin{table*}[th]
    \centering
    \begin{tabularx}{\linewidth}{lrX}
    \hline
    \textbf{Dataset}  & \textbf{Size}  & \textbf{Example}  \\
    \hline
    \textsc{WebSplit} & 938   & \textsl{Addiction journal is about addiction and is published by Wiley-Blackwell which has John Wiley \& Sons as the parent company .}    \\
    \textsc{Orlando}  & 1,104 & \textsl{She covers several British trials on sexual matters and on what might be described as trumped-up evidence: the prosecution of Penguin Books for publishing Lawrence's Lady Chatterley's Lover, 1960, the trial of ex- Liberal Party leader Jeremy Thorpe for conspiracy to murder, and the trial of Stephen Ward (described by the Oxford Dictionary of National Biography both as osteopath and scapegoat and as the British Dreyfus) for living on immoral earnings in the wake of the resignation of Minister John Profumo on 4 June 1963.} \\
    \hline
    \end{tabularx}
    \caption{Summary the two datasets of complex sentences, where \textsc{WebSplit} is synthesized and unnatural while \textsc{Orlando} contains natural sentences of \emph{absurd} complexity similar to the examples.}
    \label{tab:datasets}
\end{table*}
\section{Task Setting}
\label{sec:task}

In our studies, we consider only the hard cases of syntactic simplification \cite{niklaus-etal-2019-minwikisplit} where a complex sentence needs to be simplified into multiple ones (typically more than two).
To the best of our knowledge, there is a lack of high-quality benchmarking datasets for this task.
Synthetic and mined datasets such as WikiSplit \cite{botha-etal-2018-learning} and BiSECT \cite{kim-etal-2021-bisect} come with reference simplifications, but they only focus on binary splits, with WebSplit \cite{narayan-etal-2017-split} being an exception.
The manually labeled DeSSE dataset \cite{gao-etal-2021-abcd} is in the domain of student essays where the sentences are relatively simple.
The usefulness of the provided reference simplifications is limited, as they are often not of high quality and the granularity of the splits is pre-defined by the dataset generation process.
This motivates us to use reference-less evaluation metrics to assess the quality of the generated splits from the aspects of simplicity and meaning preservation separately \cite{cripwell2024evaluating}, and create a natural and realistic dataset of complex sentences.

\paragraph*{Datasets.}
As an instance of traditionally used benchmark datasets, we use WebSplit's test set (\textsc{WebSplit}), with the caveat that it is unnatural.
Meanwhile, we mine for sentences with high word and entity mention counts from the Orlando bibliography corpus \cite{orlando-2023},
which results in a set of structurally-complex realistic sentences expressing rich relations, written by digital humanists (\textsc{Orlando}).
Table~\ref{tab:datasets} provides a summary of the size and nature of the two datasets.

\paragraph*{Assessing Simplicity.} 
We measure the opposite of simplicity, 
the syntactic complexity of sentences, by L2SCA \cite{lu-2010-automatic}, a widely adopted set of features that highly correlate with human judgments of syntactic complexity.
It measures 14 features from five syntactic aspects. 
For the clarity of presentation, from each aspect, we choose one feature with the highest correlation with human judgments.

\paragraph*{Assessing Meaning Preservation.}
Following recent work \cite{ponce2023split,cripwell2024evaluating}, we use 
BERTScore Recall \cite{zhang2020bertscore} computed with DeBERTa-NLI\footnote{\texttt{microsoft/deberta-xlarge-mnli} as suggested by latest BERTScore guidelines.} \cite{he2021deberta} to assess whether the meaning of the original sentence is preserved in the simplification.
We do not follow previous work relying on BLEU as its lack of semantic understanding is criticized for being particularly unsuitable for simplification tasks \cite{sulem-etal-2018-simple,alvamanchego2021unsuitability}.

%% file: amrs3.tex
\section{AMR for Rule-based Simplification}
\label{sec:amrsss}

We argue that Abstract Meaning Representation (AMR) is suitable for syntactic simplification,
as its abstraction away from surface strings and syntactic structures \cite{oepen-etal-2020-mrp} allows us to define concise and interpretable rules for simplification, and its well-developed resources for parsing and generation provide a guarantee for high-quality conversions between text and graphs.
This leads to the development of \sss, an AMR-based system for syntactic simplification that is driven by a handful of simple and interpretable rules.

\subsection{Rule Set}
\label{sec:amrsss:method}

\begin{figure}[t]
    \centering
    \includegraphics[width=\linewidth]{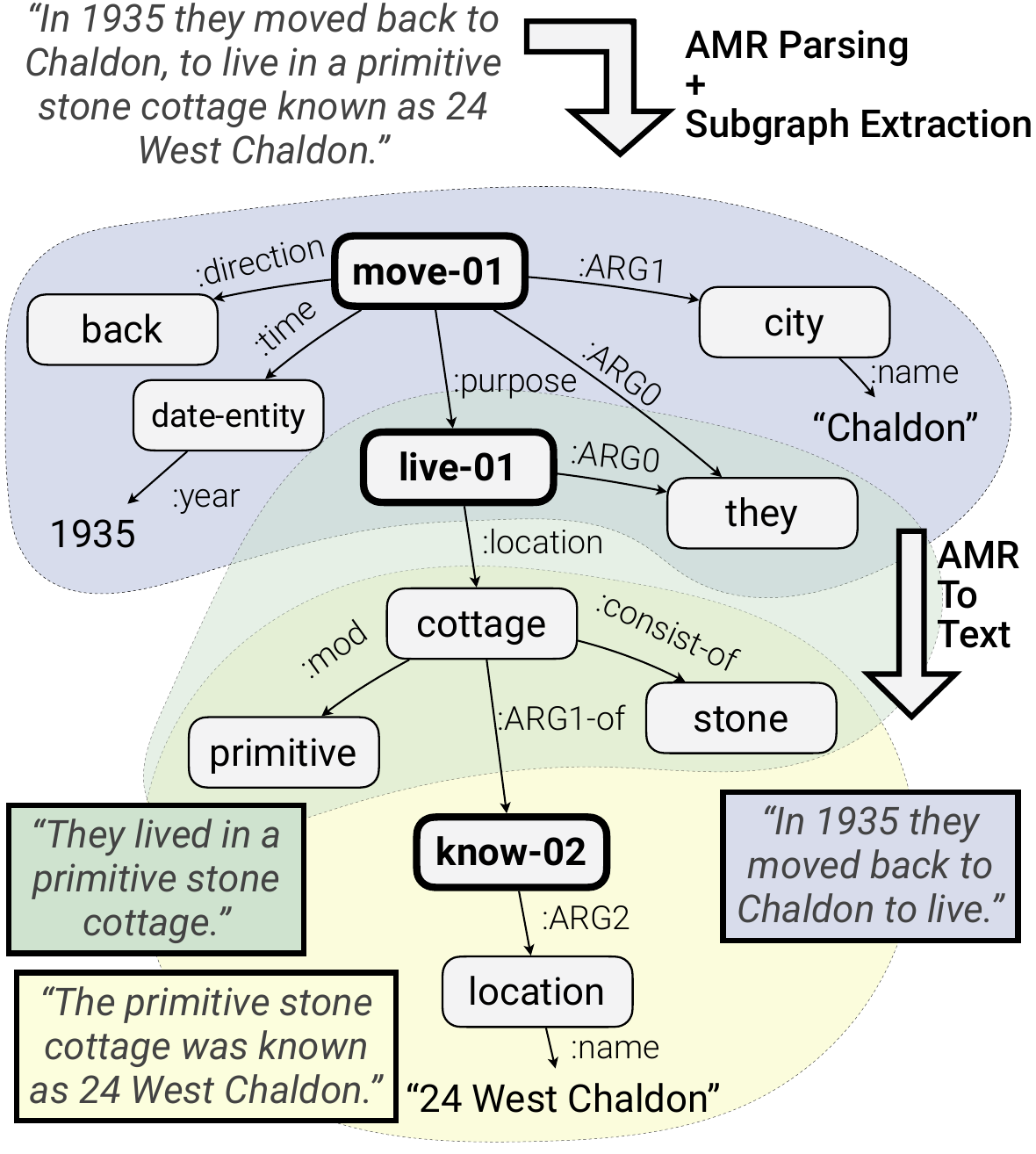}
    \caption{
        Three stages of \sss : (1) Complex input sentence (top) is parsed into an AMR graph.
        In the AMR graph, core concepts are highlighted.
        (2) Subgraphs (three encircled graphs) that correspond to simpler sentences are identified using the subgraph extraction algorithm.
        (3) The subgraphs are realized into text (three boxes at the bottom) using an AMR-to-text model.
        }
    \label{fig:amrs3}
\end{figure}
\input{alg.tex}

As illustrated in Figure~\ref{fig:amrs3},
\sss at a high-level projects a complex sentence to the space of AMR graphs using a semantic parser, and then breaks down the AMR graph of a complex sentence into a set of subgraphs, each corresponding to a semantic unit, which are then realized into simpler sentences using an AMR-to-text model.

An AMR graph (as in Fig.~\ref{fig:amrs3}) is a rooted directed acyclic graph where nodes represent \textit{concepts} and edges represent \textit{relations} between concepts \cite{banarescu-etal-2013-abstract}.
Non-leaf nodes in AMR are usually \textit{core concepts} (highlighted nodes in Fig.~\ref{fig:amrs3}) that map to predicates in OntoNotes \cite{pradhan2007ontonotes} semantic roles, and the remaining nodes are arguments of the core concepts such as (named) entities.
AMR concepts are not anchored to words, and a core concept captures an event even if the word that realizes it is a noun, adjective, or is of another part-of-speech.
This allows us to simplify the sentence by focusing on and only on the core concepts and their arguments:

\noindent\textbf{Rule 1 (Core Concept):} If a node is a core concept and has more than $\sigma$ arguments, it is considered a semantic unit, and the subgraph rooted at this node is extracted as a subgraph.

A single concept (e.g. \emph{they} in Fig.~\ref{fig:amrs3}) can be the argument of multiple core concepts. To avoid redundancy, we only extract all relations of a concept on the first occurrence and only keep non-core relations (names, values, etc. as opposed to subjects and objects) on subsequent occurrences.

\noindent\textbf{Rule 2 (Revisit):} If a node has been visited before, only extract non-core relations.

AMR by default is rooted at a single predicate (e.g. \emph{move-01}) as its focus.
Non-focused predicates, except for the arguments of the focused predicate, are connected by inverse relations (e.g. \emph{know-02} \texttt{:ARG1-of} \emph{cottage} in Fig.~\ref{fig:amrs3}) that are often realized as relative clauses.
Depending on the granularity of simplification, we may choose to extract unfocused concepts as their own subgraphs as well by reversing the direction of the inverse relations and creating a new root.

\noindent\textbf{Rule 3 (Inverse Relations):} (Optional) If a node is connected by an inverse relation, reverse the direction of the inverse relation and extract the subgraph rooted at the node.

\subsection{Implementation}

Using depth-first search (DFS) with the rules above, we extract a set of subgraphs from the AMR graph (Algorithm~\ref{alg:subgraph}), where $\sigma$ is heuristically set at 2.
We use AMRBART\footnote{\texttt{AMRBART-large-v2 (AMR3.0)}} \cite{bai2022amrbart}, a unified model with strong performance in both AMR parsing and AMR-to-text, to parse the complex sentence into AMR graphs and realize the subgraphs into text.
During AMR-to-text generation, we adopt the common practice of anonymizing named entities \cite{konstas-etal-2017-neural}.

As suggested by \citet{bai2022amrbart}, text-AMR pairs generated by semantic parsers (silver data) can benefit the training of AMR-to-text models.
To adapt AMRBART to simple sentences, we leverage this property and finetune AMRBART on silver text-AMR pairs by parsing sentences from Simple English Wikipedia\footnote{Sentences extracted from \texttt{simplewiki-20230101} dump, with 5,000 held out as test set.} using AMRBART.
After finetuning, AMRBART realizations on the held-out set achieve a BLEU of 46.23, compared to the base model's BLEU of 39.53.

\subsection{Baselines}
\input{tab-compare.tex}

We perform a comparison between \sss and the following existing systems for syntactic simplification using 
the evaluation methods outlined in \S\ref{sec:task}.
The results are reported in Table~\ref{tab:results}.

\paragraph*{DisSim.} DisSim \cite{niklaus2023dissim} performs a recursive transformation of a sentence based on a set of 35 hand-crafted syntactic and lexical rules related to the sentence’s phrase structure.

\paragraph*{ABCD.} ABCD \cite{gao-etal-2021-abcd} represents a sentence as a graph where edges are dependency and neighboring relations, and trains a neural network to predict actions on the edges.
We use its MinWiki-MLP release.

\paragraph*{DSS.} DSS \cite{sulem-etal-2018-simple} uses UCCA as the semantic representation, splits the UCCA graph based on parallel and elaborator scenes, and converts the subgraphs into text using a neural model.

\paragraph*{LLM.} We directly instruct GPT-3.5 (turbo-0125; \citealp{ouyang2022instruct}) and Llama-3 (8B-Instruct; \citealp{llama3modelcard}) with Prompt~\ref{prompt:direct}.

\vspace{1ex}
\begin{prompt}[title={Prompt \thetcbcounter: Direct Prompting}, label=prompt:direct]
    \texttt{[System]} You are a helpful assistant that simplifies syntactic structures. \\
    \texttt{[User]} Rewrite the following paragraph using simple sentence structures and no clauses or conjunctions: \texttt{\{complex sentence\}}
\end{prompt}

\subsection{Discussion}
\paragraph*{\sss achieves competitive performance without specialized supervised training.}
Overall, simplifications generated by \sss are on par with or better than the baselines in terms of meaning preservation on both datasets, as shown by the comparisons in Table~\ref{tab:results}, despite not being trained on task-specific supervised data.
The performance is close to Llama-3, a state-of-the-art LLM.
The syntactic simplicity of the generated sentences, measured by L2SCA, is at the same level as the best-performing baselines on \textsc{WebSplit} and better on \textsc{Orlando}, suggesting that the good performance of meaning preservation is not achieved by sacrificing syntactic simplicity.
The interpretable rule set of \sss makes the method easily customizable.
The comparison between \sss with and without Rule 3 exemplifies how a compromise between simplicity and meaning preservation can be made by simple adjustments of the rules.

\paragraph*{\sss enjoys unique merits beyond empirical performance.}
Specially trained models such as ABCD suffer from the lack of generalizability to new domains, as seen in its drastic performance drop on \textsc{Orlando} texts.
In contrast, AMR models that \sss relies on are trained on a diverse set of data and can be easily improved for new domains by finetuning on silver data.
LLMs are powerful and training-free, while \sss is lightweight and performs similarly well to open-weight LLMs. Admittedly, rule-based DisSim is lightweight and is performant in the evaluation.
Compared to models based on semantic representation, DisSim requires a complex set of 35 lexical and syntactic rules, while \sss only needs three simple rules. 
The rules of DisSim are crafted for English only and are hard to transfer to other languages, while despite AMR not being an interlingua \cite{banarescu-etal-2013-abstract} the rules of \sss are language-agnostic and can be easily adapted to other languages with AMR parsers.
Methods based on other semantic representations, such as UCCA-based DSS, perform worse despite having a similar workflow to \sss, showcasing the \emph{"free upgrades"} that advances in semantic representation tools can bring.

\paragraph*{Takeaways.} As \sss demonstrates, semantic graphs like AMR are mature enough to support the easy development of lightweight and interpretable systems, that still have certain advantages in LLM's age, for tasks like syntactic simplification.

%% file: alg.tex
\begin{algorithm}[t]
    \caption{Extract subgraphs from an AMR graph $G$ by performing DFS and applying the rules defined in \S\ref{sec:amrsss:method}.}\label{alg:subgraph}
    \begin{algorithmic}[1]
    \Procedure{SubGraphs}{$G$}
        \State $r \gets \varnothing$; $q\gets \{G.root\}$
        \ForAll{$e \in G.edges$, $e$ is inverse}
            \State $q\gets q \cup \{e.from\}$ \Comment{Rule 3}
            \State $e.from, e.to \gets e.to, e.from$
        \EndFor\label{alg:subgraph:inverse}
        \While{$|q| > 0$}
        \Comment{Extract from roots}
            \State $g' \gets \textsc{DFSCopy}(q.pop(), q)$
            \State $r \gets r \cup \{g'\}$
        \EndWhile\label{extract}
        \State \textbf{return} $r$
    \EndProcedure
    \Procedure{DFSCopy}{$n$, $q$}
        \State \textbf{if} $n$ is leaf \textbf{return} $n$
        \If {$n$ is core concept, $|n.edges| > \sigma$}
            \State $q \gets q \cup \{n\}$ \Comment{Rule 1}
            \State \textbf{return} $n$
        \EndIf
        \If {$n$ was visited}\Comment{Rule 2}
        \ForAll{$e \in n.edges$, $e$ is non-core}
            \State $n.addEdge(\textsc{DFSCopy}(e.to, q))$
        \EndFor
        \EndIf
        \State \textbf{else for all} {$e \in n.edges$} \textbf{do}
            \State \hspace{30pt}  $n.addEdge(\textsc{DFSCopy}(e.to, q))$
        \State \textbf{return} $n$
    \EndProcedure
    \end{algorithmic}
\end{algorithm}

%% file: tab-compare.tex
\begin{table*}[t]
    \centering
    \begin{tabular}{llllllll}
    \hline
    \multicolumn{1}{l|}{\multirow{2}{*}{\textbf{Method}}} & \multicolumn{2}{c|}{\textbf{BERTScore} $\uparrow$} & \multicolumn{5}{c}{\textbf{L2SCA} $\downarrow$} \\
    \multicolumn{1}{l|}{}                                 & \textbf{Mean}    & \multicolumn{1}{l|}{\textbf{Median}}   & \textbf{MLT}    & \textbf{C/S}  & \textbf{C/T}  & \textbf{T/S}  & \textbf{CN/T} \\ \hline
    
    \multicolumn{8}{c}{on \textsc{Orlando}}\\ \hline
    
    \multicolumn{1}{l|}{\sss}                           & 0.73    & \multicolumn{1}{l|}{0.72}     & 12.00  & 1.02 & 1.07 & 0.96 & 1.22 \\
    \multicolumn{1}{l|}{\sss (w/o Rule 3)}              & 0.79    & \multicolumn{1}{l|}{0.79}     & 18.17  & 1.30 & 1.32 & 0.98 & 1.89 \\
    \hdashline[1pt/1pt]
    \multicolumn{1}{l|}{ABCD}                             & 0.50    & \multicolumn{1}{l|}{0.51}     & 14.99  & 0.94 & 1.19 & 0.80 & 1.98 \\
    \multicolumn{1}{l|}{DisSim}                           & 0.74    & \multicolumn{1}{l|}{0.74}     & 11.15  & 1.18 & 1.16 & 1.02 & 1.24 \\
    \multicolumn{1}{l|}{DSS$^\dagger$}                           & -    & \multicolumn{1}{l|}{-}     & -  & - & - & - & - \\
    \multicolumn{1}{l|}{GPT-3.5}                          & 0.80    & \multicolumn{1}{l|}{0.82}     & 12.65  & 1.14 & 1.13 & 1.01 & 1.26 \\
    \multicolumn{1}{l|}{Llama-3-8B}                       & 0.74    & \multicolumn{1}{l|}{0.74}     & 7.89   & 1.07 & 1.07 & 1.00 & 0.70 \\ 
    \hdashline[1pt/1pt]
    \multicolumn{1}{l|}{Exact Copy}                       & 1.00    & \multicolumn{1}{l|}{1.00}     & 157.25   & 2.66 & 2.18 & 1.22 & 4.69 \\ \hline
    
    \multicolumn{8}{c}{on \textsc{WebSplit}}\\ \hline
    
    \multicolumn{1}{l|}{\sss}                           & 0.81    & \multicolumn{1}{l|}{0.81}     & 8.92   & 1.00 & 1.02 & 0.99 & 0.68 \\
    \multicolumn{1}{l|}{\sss (w/o Rule 3)}              & 0.86    & \multicolumn{1}{l|}{0.86}     & 12.26  & 1.16 & 1.16 & 1.00 & 1.16 \\
    \hdashline[1pt/1pt]

    \multicolumn{1}{l|}{ABCD}                             & 0.90    & \multicolumn{1}{l|}{0.91}     & 9.53   & 1.00 & 1.10 & 0.91 & 0.94 \\
    \multicolumn{1}{l|}{DisSim}                           & 0.87    & \multicolumn{1}{l|}{0.87}     & 8.54   & 1.05 & 1.05 & 0.99 & 0.67 \\
    \multicolumn{1}{l|}{DSS$^\dagger$}                              & 0.74    & \multicolumn{1}{l|}{0.74}     & 10.69  & 0.97 & 1.19 & 0.81 & 1.05 \\
    \multicolumn{1}{l|}{GPT-3.5}                          & 0.90    & \multicolumn{1}{l|}{0.90}     & 7.79   & 1.02 & 1.02 & 1.00 & 0.52 \\
    \multicolumn{1}{l|}{Llama-3-8B}                       & 0.84    & \multicolumn{1}{l|}{0.85}     & 6.69   & 1.01 & 1.01 & 1.00 & 0.38 \\ 
    \hdashline[1pt/1pt]
    \multicolumn{1}{l|}{Exact Copy}                       & 1.00    & \multicolumn{1}{l|}{1.00}     & 16.57   & 1.64 & 1.50 & 1.10 & 1.72 \\ \hline
    \end{tabular}
    \caption{
        \label{tab:results}
        Evaluation results of \sss and baselines on \textsc{Orlando} and \textsc{WebSplit}.
        BERTScore measures meaning preservation ($\uparrow$ the higher the better), and L2SCA measures syntactic complexity ($\downarrow$ the lower the better).
        $\dagger$ We use the output provided by \citet{sulem-etal-2018-simple} on WebSplit only, as no code is available.
        Five L2SCA metrics correspond to production unit length, overall complexity, subordination, coordination, and phrasal complexity.
        See \citet{lu-2010-automatic} for the exact definition of L2SCA metrics.
    }
\end{table*}

%% file: prompting.tex
\section{AMR for LLM-based Simplification}
\label{sec:prompting}

\input{tab-prompting.tex}

Given the position of AMR as an expressive and suitable intermediate for syntactic simplification and LLM's strong performance in the task, a natural question arises as to whether AMR can be used as an auxiliary to LLMs to improve their performance in syntactic simplification in the scenario of zero-shot prompting.
We investigate this question by designing a set of controlled prompting strategies to examine how the elements of AMR affect LLM.
This is an addition to \citet{jin2024analyzing} which tested directly appending AMR to the prompt in a variety of tasks, while syntactic simplification was not included in their study.
Extending their work, we explore a new prompting strategy (named AMR Chain-of-Code or AMRCoC) that guides LLMs to perform explicit symbolic reasoning over AMR graphs instead of making implicit inferences as in \citet{jin2024analyzing}.

\begin{figure}[t]
\begin{prompt}[title={Prompt \thetcbcounter: Direct Full AMR Prompting (Jin et al., 2024)}, label=prompt:direct-amr]
    \texttt{[User]}   You are given a paragraph and its abstract meaning representation (AMR).\\
    \# Paragraph\\
    \texttt{\{complex sentence\}}\\
    \# AMR\\
    \texttt{\{amr\}}\\
    Rewrite the paragraph using simple sentence structures and no clauses or conjunctions.
    You can refer to the provided AMR if it helps you in the rewriting.\\
    The rewritten paragraph:
\end{prompt}
\end{figure}

\subsection{Direct AMR Prompting}
\citeposs{jin2024analyzing} evaluation framework simply supplies linearized AMR in PENMAN format \cite{Matthiessen1992penman} in parallel with text, providing only vague instructions to the LLM on how to use the AMR, and requiring the LLM to directly produce the output \emph{without} \footnote{Despite having an imprecise name "AMR for Chain-of-Thought" prompting in the original paper.} explicitly producing reasoning steps.
To add to their tests, we adapt their framework to the syntactic simplification task as in Prompt~\ref{prompt:direct-amr}.

\paragraph*{Performance.} Interestingly, our evaluations (Table~\ref{tab:prompting}) show that the direct AMR prompting does not harm the performance of LLMs in syntactic simplification, and in some cases, it provides improvements especially for more complex inputs.
This adds syntactic simplification as a counterexample to the findings of \citet{jin2024analyzing}.

\paragraph*{Effect of Elements.}
To isolate the effects of different elements (subgraphs, entities, and predicates) of AMR, we further design a set of controlled prompts following the same format of Prompt~\ref{prompt:direct-amr}, where the linearization of complete AMR is replaced by specific parts of the AMR: 

\noindent (1) Instead of the sole AMR corresponding to the whole complex sentence, we provide a list of AMR graphs extracted with Algorithm~\ref{alg:subgraph} for each semantic unit in the sentence (\textbf{subgrpahs});

\noindent (2) We provide only a list of predicates in the AMR (\textbf{predicates}), e.g. \emph{"move, live, know"} as in Figure~\ref{fig:amrs3};

\noindent (3) We provide only a list of entities as reflected by the non-core concepts in the AMR (\textbf{entities}), e.g. \emph{"date (1935), they, city (Chaldon), location (24 West Chaldon)"} as in Figure~\ref{fig:amrs3}.

Both predicates and entities provide incomplete information about the events of a sentence, while not requiring LLM's capability to reason over a symbolic graph.
However, we find that for the tasks and LLMs in question, LLMs are capable of directly and implicitly using information in the AMR as appropriate, while trading information completeness for the ease of symbolic graph processing offers more harms than benefits (Table~\ref{tab:prompting}).

\paragraph*{Takeaways.} Directly supplying AMR to LLMs is not monochromatically harmful across tasks.
Growing the list of tasks benefited and harmed by direct AMR prompting is needed to draw conclusions on the role of meaning representations in the LLM era.

\subsection{AMRCoC Prompting}

\begin{figure}[t]
    \begin{prompt}[title={Prompt \thetcbcounter: AMR-Chain-of-Code Prompting (with simplification and an example)}, label=prompt:chain-of-amr]
        \texttt{[User]}\\
        You are given a paragraph and its abstract meaning representation (AMR).
        AMR captures ``who is doing what to whom'' in a sentence. Begin your steps with \# Steps.\\
        \textbf{\# Functions to process AMR}\\
        \texttt{EXTRACT\_SUBGRAPH(root: Node) -> AMR} \\ extracts the subgraph rooted at `root`.\\
        \texttt{AMR\_TO\_TEXT(graph: AMR) -> str} \\ converts an AMR graph back to text.\\
        \textbf{\# Example program}\\
        \texttt{for predicate in amr:}\\
        \texttt{g = EXTRACT\_SUBGRAPH(predicate)}\\
        \texttt{print(AMR\_TO\_TEXT(g))}\\
        \\
        \texttt{[Assistant]}\\
        \textbf{\# Steps}\\
        \texttt{g = EXTRACT\_SUBGRAPH(flow-01) => (z1 / flow-01 :ARG1(...) ...)}\\
        \texttt{print(AMR\_TO\_TEXT(g))}\\
        \texttt{=> It flows through the town of Yeovil.}\\
        \texttt{g = EXTRACT\_SUBGRAPH(join-01) => (z5 / join-01
        :ARG1 (z2 / it) ...)}\\
        \texttt{...}\\
        \textbf{\# Output}\\ It flows through the town of Yeovil. It joins River Parrett.
    \end{prompt}
    \end{figure}

Despite the evidence that LLMs can benefit from direct AMR prompting, it is widely accepted (\citealp{wei2022cot,saparov2023greedy}, \emph{inter alia}) that LLM's reasoning capability over complex tasks (e.g. processing AMR) can be improved by explicitly decomposing them into reasoning steps. 
To remedy the lack of explicit reasoning, we build on Chain-of-Code (CoC) prompting \cite{li2023chainofcode}, where pseudocode execution is shown helpful for the LLM to perform explicit algorithmic reasoning in general tasks, and design AMRCoC prompting (Prompt~\ref{prompt:chain-of-amr}): LLM is guided to produce explicit reasoning steps over AMR graphs by using functions to process AMR, and an example program that demonstrates the use of these functions.
The functions and programs are not formally defined but in the form of function signatures or pseudocode, as we expect LLM to emulate the execution \cite{li2023chainofcode,chae2024llmcompiler}.

\paragraph*{Performance.}
AMRCoC offers the same level of meaning preservation (last rows of Table~\ref{tab:prompting}) compared to direct AMR prompting, although the simplicity of generations degrades to the level of \sss, which is perhaps unsurprising as we prompt the LLM to follow a similar algorithm.
The example program in the prompt may not be optimal, but it is possible to synthesize or improve the program using LLM \cite{chae2024llmcompiler}.

\paragraph*{Emulated Execution.}
More importantly, the breakdown of AMRCoC execution (Table~\ref{tab:coc}) verifies that LLMs can be prompted to perform explicit algorithmic reasoning over AMR graphs, which is a promising direction for future research.
LLM almost always emulates the execution of the example pseudocode program ("Following algorithm" in Table~\ref{tab:coc}).
The extracted AMR graphs, although not always grammatically correct especially for complex inputs ("Grammatical AMR"), are not hallucinated and are based on existing nodes and edges in the input AMR ("Node and edge existence"), and mostly match the real execution results of Algorithm~\ref{alg:subgraph} ("Matching algorithm output"). When combined, AMR graphs extracted by LLM cover most of the semantic information in the input AMR ("Node coverage"), providing a guarantee for meaning preservation.

\paragraph*{Takeaways.} Chain-of-Code prompting provides a way for LLM to perform symbolic reasoning over semantic graphs via algorithm emulation. This provides a way to bring algorithmic graph processing to LLMs for semantic-centered NLP applications, to enjoy the benefits of both worlds.

\input{tab-coc.tex}

%% file: tab-prompting.tex
\begin{table*}[!t]
    \centering
    \resizebox{0.95\textwidth}{!}{
    \begin{tabular}{llllllllll}
    \hline
    \multirow{2}{*}{} & \multirow{2}{*}{\textbf{Prompting}} & \multicolumn{2}{c}{\textbf{BERTScore} $\uparrow$} & \multicolumn{1}{c|}{} & \multirow{2}{*}{} & \multirow{2}{*}{\textbf{Prompting}} & \multicolumn{2}{c}{\textbf{BERTScore} $\uparrow$} &   \\
                                    &                                     & \textbf{Mean}     & \textbf{Median}    & \multicolumn{1}{l|}{\textbf{MLT} $\downarrow$}   &                                 &                                     & \textbf{Mean}     & \textbf{Median}    & \textbf{MLT} $\downarrow$   \\ \hline
    
    \multicolumn{10}{c}{on \textsc{Orlando}} \\ \hline
    
    GPT-3.5                         & \textit{vanilla}                    & 0.80              & 0.82               & \multicolumn{1}{l|}{12.65}          & Llama-3                      & \textit{vanilla}                    & 0.74              & 0.74               & 7.89           \\
                                    & \textit{direct AMR}                 & 0.81              & 0.82               & \multicolumn{1}{l|}{12.79}          &                                 & \textit{direct AMR}                 & 0.78              & 0.78               & 11.74          \\
                                    & \textit{subgraphs}                  & 0.80              & 0.81               & \multicolumn{1}{l|}{11.65}          &                                 & \textit{subgraphs}                  & 0.78              & 0.78               & 12.45          \\
                                    & \textit{entities}                   & 0.79              & 0.80               & \multicolumn{1}{l|}{10.99}          &                                 & \textit{entities}                   & 0.70              & 0.71               & 7.75           \\
                                    & \textit{predicates}                      & 0.73              & 0.74               & \multicolumn{1}{l|}{7.34}           &                                 & \textit{predicates}                      & 0.70              & 0.70               & 7.55           \\
                                    & \textit{AMRCoC}                    & 0.79              & 0.81               & \multicolumn{1}{l|}{17.29}          &                                 & \textit{AMRCoC}                   & 0.76              & 0.77               & 14.03          \\ \hline

    \multicolumn{10}{c}{on \textsc{WebSplit}} \\ \hline

    GPT-3.5                         & \textit{vanilla}                    & 0.90              & 0.90               & \multicolumn{1}{l|}{7.79}           & Llama-3                     & \textit{vanilla}                    & 0.84              & 0.85               & 6.69           \\
                                    & \textit{direct AMR}                 & 0.88              & 0.89               & \multicolumn{1}{l|}{8.59}           &                                 & \textit{direct AMR}                 & 0.83              & 0.85               & 8.15           \\
                                    & \textit{subgraphs}                  & 0.87              & 0.88               & \multicolumn{1}{l|}{8.35}           &                                 & \textit{subgraphs}                  & 0.82              & 0.84               & 7.41           \\
                                    & \textit{}                           &                   &                    & \multicolumn{1}{l|}{}               &                                 & \textit{entities}                   & 0.78              & 0.79               & 6.63           \\
                                    & \textit{}                           &                   &                    & \multicolumn{1}{l|}{}               &                                 & \textit{predicates}                      & 0.76              & 0.77               & 7.12           \\
                                    & \textit{AMRCoC}                    & 0.89              & 0.90               & \multicolumn{1}{l|}{9.15}           &                                 & \textit{AMRCoC}                   & 0.84              & 0.85               & 8.27           \\ \hline

    \end{tabular}
}
\caption{
    \label{tab:prompting}
    Evaluation results of GPT-3.5 and Llama-3 on \textsc{Orlando} and \textsc{WebSplit} with different prompting strategies. Notations are consistent with Table~\ref{tab:results}. Due to space limit, we only show one L2SCA metric, MLT, that has the highest variance across prompts.
}
\end{table*}

%% file: tab-coc.tex
\begin{table}[t]
    \centering
    \resizebox{\columnwidth}{!}{
    \begin{tabular}{lll}
    \hline
      \textbf{Property} & \textbf{Orlando} & \textbf{Websplit} \\ \hline
    Following algorithm       & 99.8\%           & 92.8\%            \\
    Grammatical AMR           & 31.3\%           & 67.8\%            \\
    Node and edge existence   & 98.6\%           & 99.7\%            \\
    Node coverage             & 72.3\%           & 90.0\%            \\
    \hdashline[1pt/1pt]
    Matching algorithm output & 52.1\%           & 66.0\%            \\ \hline
    \end{tabular}
    }
    \caption{
        \label{tab:coc}
        Success rates of Llama-3's Chain-of-Code execution at different stages. Numbers are macro-averaged across all input complex sentences. For the first four rows, higher values are always favored.
    }
\end{table}

%% file: misc.tex
\section{Conclusion}
In light of recent developments in semantic representations and LLMs, we presented a retrospective view of using semantic representation graphs for syntactic simplification, with refreshed datasets and up-to-date semantic representation models. 
In prospect, we added to the case studies of the beneficial and harmful effects of using AMR for LLM, and proposed a new AMRCoC prompting strategy with the potential of bridging symbolic and graphical algorithms to LLMs.

\section*{Limitations}
The proposed \sss is not the best performing syntactic simplification system in terms of having the highest absolute numbers of BERTScore and L2SCA metrics across the datasets, as is particularly overshadowed by LLMs.
The main conclusion is more about the current state of semantic representations: they are still handy in building solutions for semantic tasks, and that solution can have merits that make it a good fit in certain scenarios.
Despite that, the design of AMR has some disadvantages that make it less effective to be used out-of-the-box for text simplification, namely the absence of inflectional morphology for tense and number. \citet{banarescu-etal-2013-abstract} suggested that this can be remedied by adding these notions to AMR as an extension, which is a direction for future work.

Our evaluation of syntactic simplification is limited to automated methods.
Although previous work has shown high correlations between the metrics we use and human judgments on meaning preservation, syntactic complexity, and reading difficulty, we acknowledge that those conclusions might not hold for domains out of their respective evaluations.
A systematic evaluation method, tailored to the specific task of syntactic simplification and aligned with human judgments, similar to \citet{alvamanchego2021unsuitability,maddela-etal-2023-lens}, would be beneficial for similar studies but is out of the scope of this work.

Finally, the applicability of AMRCoC prompting is only tested on the single task of syntactic simplification. Although the properties it demonstrates are promising, we have yet to test it on other tasks such as the ones in \citet{jin2024analyzing}.

\section*{Acknowledgements}
We acknowledge the support of the Natural Sciences and Engineering Research Council of Canada (NSERC). This work is also supported in part by a gift from Scotiabank. We thank Natalie Hervieux and Dr. Susan Brown for preparing the \textsc{Orlando} dataset. We also thank the anonymous reviewers for their valuable feedback.